\PassOptionsToPackage{pdftex}{graphicx}
\documentclass{article} 
\usepackage{iclr2027_conference,times}
\usepackage{booktabs}
\usepackage{multirow}
\usepackage{graphicx}

\usepackage{amsmath,amsfonts,bm}

\def\eqref#1{equation~\ref{#1}}

\def\1{\bm{1}}

\DeclareMathAlphabet{\mathsfit}{\encodingdefault}{\sfdefault}{m}{sl}
\SetMathAlphabet{\mathsfit}{bold}{\encodingdefault}{\sfdefault}{bx}{n}

\usepackage{multirow}
\usepackage{wrapfig}
\usepackage{graphicx}
\usepackage{xspace}
\usepackage{booktabs}
\usepackage{enumitem}
\usepackage{nameref}
\usepackage{array}
\usepackage{hyperref}
\usepackage{url}
\usepackage{ulem}
\usepackage{bm}
\usepackage{amsmath,amssymb}
\usepackage{subcaption}
\usepackage{booktabs}
\usepackage{setspace}
\usepackage{makecell}
\usepackage{wrapfig}
\usepackage{tabularx}
\usepackage{xcolor} %
\hypersetup{
    colorlinks,
    linkcolor={red!50!black},
    citecolor={brown!85!red},
    urlcolor={blue!80!black}
}

\newcommand{\gcheck}{\textcolor{green!60!black}{\checkmark}}
\newcommand{\na}{--}

\title{
RL-PaO: Prediction as Action in \\Decision Making under Uncertainty
}

\author{
    Jiahui Feng\textsuperscript{$\spadesuit$},
    Dafang Zhao\textsuperscript{$\lozenge$},
    Zheng Chen\textsuperscript{$\clubsuit$},
    Zhengmao Li\textsuperscript{$\heartsuit$},
    Lingwei Zhu\textsuperscript{$\spadesuit\dagger$} \\[3pt]
    \textsuperscript{$\spadesuit$} Great Bay University, School of Computing and Information Technology, China \\
    \textsuperscript{$\lozenge$} The University of Osaka, School of Information Science and Technology, Japan \\
    \textsuperscript{$\clubsuit$} The University of Osaka, SANKEN, Japan \\
    \textsuperscript{$\heartsuit$} Aalto University, School of Electrical Engineering, Finland \\[4pt]
    \textsuperscript{$\dagger$}Corresponding author: \texttt{zhulingwei@gbu.edu.cn}
}

\begin{document}

\makeatletter
\iclrfinaltrue
\makeatother

\maketitle

\makeatletter
\fancyhead{}
\lhead{}
\chead{}
\rhead{}
\renewcommand{\headrulewidth}{0.4pt}
\thispagestyle{fancy}
\makeatother
\begin{abstract}

Decision-making under uncertainty often relies on predicted parameters, yet accurate prediction does not necessarily lead to good operational decisions. Aligning prediction with downstream optimization requires learning from the consequences of the decisions those predictions induce. 
We introduce RL-PaO, a reinforcement learning framework that  integrates system formulation, optimization, and decision execution into a single environment. 
This yields a Markov decision process in which prediction is regarded as action: it shifts the environment to produce subsequent context and reward that explicitly aligns  prediction error with realized cost, and learning the optimal policy does not require differentiating through the black-box solver. 
We evaluate RL-PaO on day-ahead energy scheduling using real historical data. 
On the test year, RL-PaO achieves the lowest annual cost among the non-oracle baselines, achieving on average $10\%$ cost reduction. 
Moreover, RL-PaO is capable of further analyses to provide strong interpretability both from the policy evolution perspective and  the cost-accuracy trade-off.

\end{abstract}

\section{Introduction}

Decision-making under uncertainty is fundamental to energy systems, transportation, and supply chains \citep{mandi2024decision}. Operational decisions must be made before parameters or constraints become known. Predictive models estimate these parameters, optimization solvers generate decisions based on the estimates and execution under realized conditions determines their value. 
It is vital to evaluate prediction quality since it changes the optimization objective and influences downstream execution.
It remains a central challenge to align the predictor with the operational purpose.

\emph{Predict-then-optimize} (PtO) trains the predictor independently of the optimizer, but its training objective such as MSE, MAE can be misaligned with downstream cost: lower estimation error does not necessarily translate to better decisions \citep{elmachtoub2022spo}. 
\emph{Prediction-and-optimization} (PaO) addresses this mismatch by feeding back decision-loss surrogates to augment the prediction loss  \citep{mandi2025feasibility,silvestri2026score}, in the hope that by minimizing the augmented loss, a balance between prediction accuracy and decision quality could be reached so as to improve downstream performance \citep{mandi2024decision}.
However, training on the composite loss  could lead to undesired solutions that are suboptimal for all of the member losses, see Table \ref{tab:method_comparison} for detailed comparison. 
Therefore, a paradigm that can effective align prediction and optimization is called for.

\begin{figure}[t]
        \centering
        \includegraphics[width=0.975\textwidth,trim={0.5cm 0cm 0.5cm 0cm},clip]{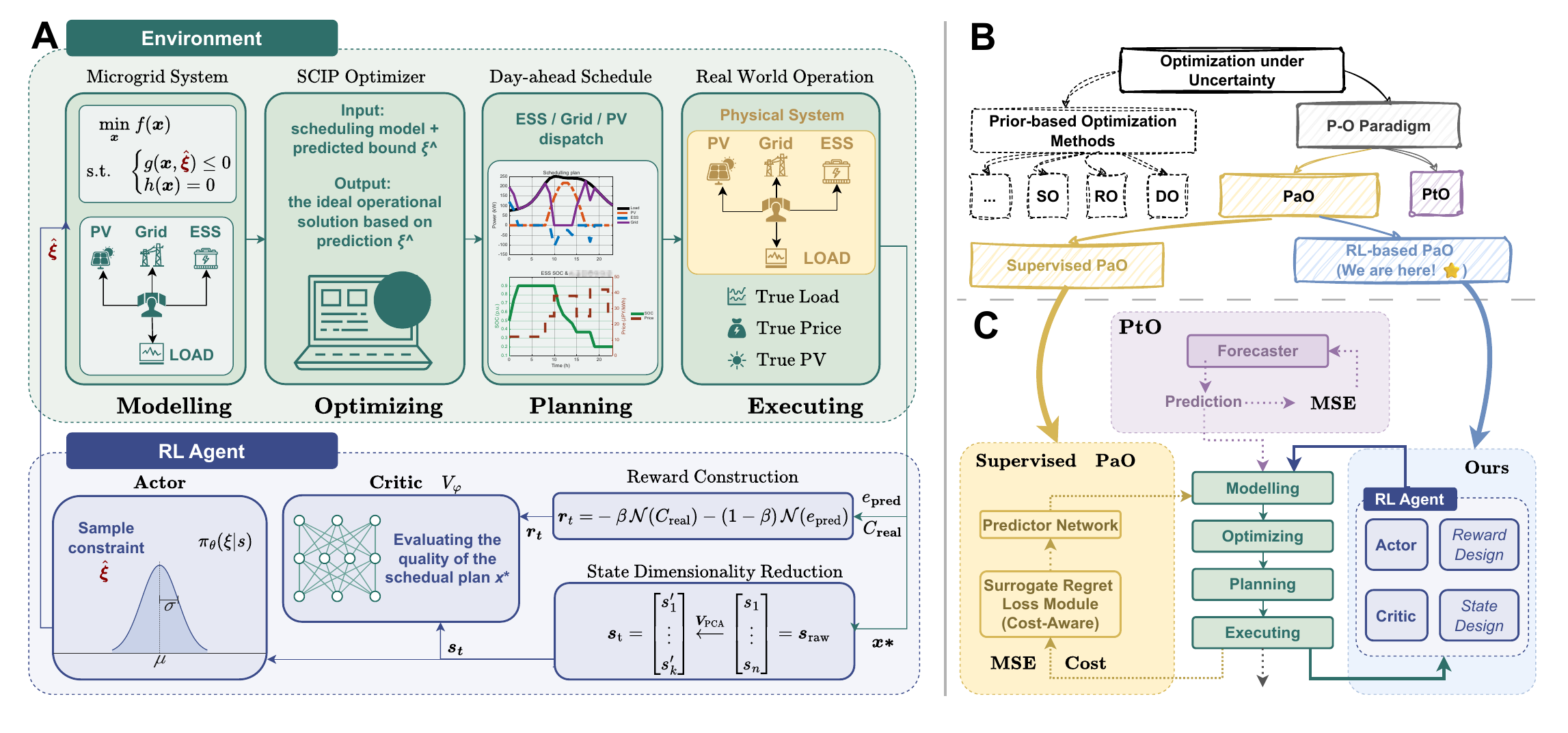}
        \vspace{-5pt}
        \caption{
        Overview of the proposed RL-PaO framework.
        \textbf{(A)} We explicitly include system modeling, optimization, scheduling and real-world operation into one environment. 
        The RL agent observes current context to predict the unknown constraint in the optimization objective to be solved by an MILP solver. 
        The solution influences execution process, yielding reward and next context.
        \textbf{(B)} our RL formulation is under the PaO category and different from the conventional supervised PaO methods.
        \textbf{(C)} Architectural comparison of different methods. PtO uses an open-loop predictor. Supervised PaO closes the training loop by incorporating downstream loss. 
        By contrast, the proposed method aligns prediction and optimization by a novel MDP and RL training.
        }
        \label{fig:framework}
    \end{figure}

In this paper, we introduce \textbf{RL-PaO}.
By treating system formulation, objective optimization and schedule execution as an integrated environment, we convert prediction into action that can shift  the environment to produce next solution and a scalar reward that explicitly evaluates the cost-accuracy trade-off for the current solution. I.e., we build a Markov Decision Process (MDP) \citep{Puterman1994} that closes the loop between the environment and an agent policy, see Figure \ref{fig:framework}.
Though RL has been utilized in improving solver procedures such as MILP \citep{Tang2020-RLforIP,Qi2021-RLforMILP}, RL-PaO differs from them in that  RL-PaO concerns alignment between prediction and blackbox optimization, not accelerating a solver conditional on its internal states.

We evaluate RL-PaO on a day-ahead scheduling problem using real history data from the Osaka University.
We train a Proximal Policy Optimization (PPO) agent \citep{schulman2017proximal} on the train set and evaluate on a held-out test year.
Results show that our method achieves the lowest annual cost among all baseline methods except the oracle.
Moreover, unlike conventional PtO and PaO, our formulation provides strong interpretability to characterize the cost--accuracy trade-off: prediction error is insufficient to assess decision quality
In short, our contributions are:
\begin{itemize}[left=0pt]
    \item \textbf{(Novel paradigm) }We demonstrate that RL -- conventionally leveraged for closed-loop system control -- can also serve as a decision-oriented prediction module. On top of this, we formalize the two-stage prediction-optimization pipeline as a Markov Decision Process, based on which stable convergence and robustness towards hyperparameters is achieved.
    \item \textbf{(Parsimonious formulation)} We propose a counterintuitive state-action construction where the state is defined exclusively by the downstream system operational schedule. We prove that the schedule plan inherently encodes sufficient information to infer the upper bound of the constraints, removing the need for extra environmental input that supervised forecasting methods rely on.
    \item \textbf{(Real-world evaluation)} We evaluate RL-PaO on a real day-ahead scheduling problem at the Osaka University. By learning from offline data, the trained RL agent outperforms all non-oracle baselines in terms of annual cost, $9.42\%$ better than supervised PaO and $12.07\%$ than PtO.
    RL-PaO also provides strong interpretability from multiple perspectives.  
    \item \textbf{(High generalizability)} Our framework treats the downstream task as a black box, and is inherently agnostic to the underlying system model, mathematical problem formulation, and prediction target. This yields strong generalizability and enables a truly model-free, problem-agnostic paradigm for decision-making problem.
\end{itemize}

\begin{table*}[t]
\centering
\resizebox{\textwidth}{!}{
\begin{tabular}{lccccccc}
\toprule
& \multicolumn{3}{c}{\textbf{Prior-based Optimization}}
& \multicolumn{3}{c}{\textbf{Predict-Optimize Paradigm}} \\
\cmidrule(lr){2-4} \cmidrule(lr){5-7}
& & & & \textbf{PtO} & \multicolumn{2}{c}{\textbf{PaO}} \\
\cmidrule(lr){6-7}
& & & & & Supervised PaO & RL-PaO \\
& DO & RO & SO &  &  & \textbf{Ours} \\
\midrule
Uncertainty modeling
& \texttimes
& \makecell{Robust \\ uncertainty sets}
& \makecell{Probability \\ distributions}
& \makecell{Open-loop \\ prediction}
& \makecell{Closed-loop \\ prediction}
& \makecell{Prediction-optimization \\ alignment}\\


Placement-free uncertainty
& \na & \gcheck & \gcheck & \gcheck & \texttimes  & \gcheck \\
Decision-focused training
& \na & \na & \na & \texttimes & \gcheck  & \gcheck \\
Model-agnostic optimization
& \na & \na & \na & \gcheck & \texttimes  & \gcheck \\
Gradient-free
& \na & \na & \na & \gcheck & \texttimes & \gcheck \\
\bottomrule
\end{tabular}
}
\caption{Comparison of methods for optimization under uncertainty. 
Ours is the first RL-based PaO method that aims at improving the alignment between prediction and optimization.
We emphasize the importance of \textit{Decision‑focused training} that optimizes by downstream decision cost instead of prediction error; and 
\textit{Model‑agnostic optimization} that operates without full knowledge of the underlying optimization formulation. 
}
\vspace{-10pt}
\label{tab:method_comparison}
\end{table*}

\section{Background and Problem Setting}
\label{sec:background}

In this paper we consider the following generic optimization problem parameterized by an uncertain quantity in vector $\bm{\xi} = \{\bm{\xi}_f,\bm{\xi}_g,\bm{\xi}_h\}$:
\begin{align}
\begin{split}
    \bm{x}^{*}(\bm{\xi})
    &=
    \arg\min_{\bm{x}\in\mathcal{X}}
    f(\bm{x},\bm{\xi}_f),\\
    \text{s.t.}\qquad
    &\bm{g}(\bm{x},\bm{\xi}_g) \leq \bm{0},\\
    &\bm{h}(\bm{x},\bm{\xi}_h) = \bm{0},
    \end{split}
\label{eq:generic_optimization} 
\end{align}
where $\bm{x}$ denotes the operational decision, $f$ is the operational objective, and $\bm{g}$ and $\bm{h}$ describe inequality and equality constraints, respectively.
In practice, the decision must be made before $\bm{\xi}$ becomes available.
Therefore, from the perspective of how the uncertainty on $\bm{\xi}$ is incorporated, existing approaches can be broadly organized into two families.
The first family leverages nominal values, uncertainty sets, or probability distributions.
In the rest of the paper,
we refer to this family collectively as \emph{Prior-based Optimization (PbO)}.
The second family exploits observable contextual information to estimate the unknown quantities and uses them in the optimization process. 
We refer to this family as \emph{Prediction-Optimization (P-O)}.~\citep{lahoud2025predict, mandi2024decision}

\textbf{Prior-based Optimization.} 
Prior-based opimization specifies a representation of uncertainty before solving the optimization problem. 
Depending on how uncertainty is represented, three representative formulations are deterministic optimization (DO), robust optimization (RO), and stochastic optimization (SO).
DO infers fixed estimate $\hat{\bm{\xi}}$ from historical data. 
SO assumes a predefined probability distribution for $\bm{\xi}$, reformulating the problem using expectations; i.e., minimizing $\mathbb{E}_{\bm{\xi}_f}[f(\bm{x}, \bm{\xi}_f)]$ subject to $\mathbb{E}_{\bm{\xi}_g}[\bm{g}(\bm{x}, \bm{\xi}_g)] \leq \bm{0}$ and $\mathbb{E}_{\bm{\xi}_h}[\bm{h}(\bm{x}, \bm{\xi}_h)] = \bm{0}$. 
RO bounds $\bm{\xi}$ within a predefined uncertainty set.
Optimizing the worst-case objective $\max_{\bm{\xi}_f} f(\bm{x}, \bm{\xi}_f)$ while ensuring constraints strictly hold for all realizations $\bm{\xi}_g$ and $\bm{\xi}_h$.
These methods demand analytical priors for $\bm{\xi}$, which are often intractable in complex environments, especially for SO. Our approach directly learns optimal strategies, circumventing the requirements on mathematical priors and explicit uncertainty modeling.

\textbf{PtO and PaO. }
Besides the prior-based methods above, 
addressing uncertainty through prediction in advance or jointly with optimization has been gaining popularity.
Let $\bm{z}$ denote a latent variable available at the decision time,  $\bm{\xi}$ an uncertain parameter required by the downstream optimization problem. 
A predictive model $q_{\bm{\theta}}$ parameterized by $\bm{\theta}$ maps the latent variable to an estimate, then fed into the objective function to construct a decision $\bm{x}^*$:
\begin{align}
\hat{\bm{\xi}} = q_{\bm{\theta}}(\bm{z}), \qquad      
\bm{x}^{*}(\hat{\bm{\xi}}) = \arg\min_{\bm{x}\in\mathcal{X}} f(\bm{x},\hat{\bm{\xi}}).
\label{eq:prediction}
\vspace{-10pt}
\end{align}
The decision $\boldsymbol{x}^*$ is then executed as a plan to obtain downstream cost.
When the prediction model is trained independently of the optimizer, the method is called \emph{pedict-then-optimize} (PtO). On the contrary,  it is called \emph{prediction-and-optimization} (PaO).
PaO explicitly feeds back decision quality to the prediction model to help improve training.

We identify the following questions that cannot be readily answered by the existing methods:
\begin{enumerate}
\item Prior-based methods require knowing uncertain characterizations, which are typically intractable in complex environments. 
\item standard PtO methods typically operate in an open-loop manner, relying on minimizing estimation errors  without feedback from realized operational outcomes \citep{chenFeatureDrivenEconomicImprovement2022, lahoud2025predict}.
However, lower estimation errors do not necessarily translate to better decisions or lower downstream costs \citep{mandi2024decision}.
\item Existing PaO methods closes the loop by feeding back downstream costs to augment the estimation error (e.g. MSE), in a hope that minimizing the augmented loss would balance minimizing cost and MSE. However, performing gradient descent on the added loss could lead to solutions that are suboptimal for all member losses. \citep{shah2022lodl}

\end{enumerate}


\textbf{Problem Formulation. }
We consider the following instance of Equation \ref{eq:generic_optimization}: 
day-ahead scheduling electricity use.
The problem focuses on uncertainty in the inequality-constrained photovoltaic (PV) generation upper bound, a scenario that is particularly challenging for existing PaO methods.
The system consists of four core components: 
(1) PV generation provides variable energy input;
(2) Energy Storage Systems (ESS) shift power consumption and supply across time periods;
(3) Load represents fixed electricity demand, and 
(4) power exchange with the main power grid. 
The problem can be cast as the Mixed Integer Linear Programming (MILP) form:
\begin{equation}
\begin{aligned}
&\min_{\boldsymbol{P^\text{grid}}} \quad\boldsymbol{c}^\top \boldsymbol{P}^\text{grid}\\
&\text{s.t. } \quad \boldsymbol{1}_{\pm}^\top\boldsymbol{P}_t  = \boldsymbol{0}, \quad  \boldsymbol{0} \leq \boldsymbol{P}_{t} \leq \boldsymbol{U}_t, 
\quad \boldsymbol{E}_t = \boldsymbol{E}_{t-1} + \left(\eta_\text{char} \cdot \boldsymbol{p}^\text{char}_{t} - \eta_\text{disc}^{-1} \cdot\boldsymbol{p}^\text{disc}_t\right), \quad \forall t\\
& \text{where} \quad \boldsymbol{P}_t := \begin{bmatrix}
    P^{\text{pv}}_t \\
    P^{\text{grid}}_t \\
    P^{\text{load}}_t \\
    \boldsymbol{p}^{\text{char}}_{t} \\
    \boldsymbol{p}^{\text{disc}}_{t}\\
\end{bmatrix},
\,\boldsymbol{U}_t :=
\begin{bmatrix}
    U^{\text{pv}}_t \\
    U^{\text{grid}}_t \\
    \infty\\
    \boldsymbol{u}_{t}\cdot U^{\text{char}}_\text{group} \\
    (1-\boldsymbol{u}_{t})\cdot U^{\text{disc}}_\text{group}\\
\end{bmatrix},
\, \boldsymbol{1}_{\pm} := \begin{bmatrix}
    +1 \\
    +1\\
    -1\\
    -\boldsymbol{1}\\
    +\boldsymbol{1}
\end{bmatrix}.
\end{aligned}
\label{eq:milp_full}
\end{equation}
\begin{wraptable}[16]{r}{0.4\textwidth}
\centering
\footnotesize
\setlength{\tabcolsep}{4pt}
\vspace{-10pt}
\caption{Notations for the problem.}
\vspace{-5pt}
\label{tab:nomenclature}
\begin{tabularx}{\linewidth}{@{}lX@{}}
\toprule
Symbol & Description \\
\midrule
$\boldsymbol{c}$
    & Electricity price \\
$P_t^{\text{pv}}$
    & Actual PV output \\
$P_t^{\text{grid}}$
    & Grid import power \\
$P_t^{\text{load}}$
    & Fixed load demand \\
$\boldsymbol{p}_t^{\cdot}$
    & ESS (dis)charging power \\
$U_t^{\text{pv}}$
    & PV upper bound$^\dagger$ \\
$U_t^{\text{grid}}$
    & Grid exchange limit \\
$U_{\text{group}}^{\cdot}$
    & (Dis)charge power limit \\
$\boldsymbol{E}_t$
    & Battery state \\
$\eta_{\cdot}$
    & (Dis)charging efficiency \\
$\boldsymbol{u}_t$
    & Binary ESS mode indicator \\
$\mathcal{M}$
    & ESS unit set \\
\bottomrule
\multicolumn{2}{@{}p{\linewidth}@{}}{%
    $^\dagger$ Uncertain parameter in the problem.
}
\end{tabularx}
\end{wraptable}
where $\boldsymbol{c}$ denotes the cost vector, 
$\boldsymbol{u}_{t}, \boldsymbol{p}_{t}^{\cdot}$  and $\boldsymbol{1}$ are $|\mathcal{M}|$-dimensional vectors, with $\mathcal{M}$ denoting the full set of ESS units.
$P^\text{pv}_{t}$ is actual PV generation, ${P}^\text{load}_{t}$ is the fixed load demand.
$\boldsymbol{p}_{t}^{\cdot}$ are the (dis)charging power.
$\boldsymbol{E}$ the state of battery that should be kept in a fixed range.
Intuitively, the objective minimizes total grid electricity purchase cost over the scheduling horizon, with $\boldsymbol{c}$ denoting the electricity price, $P^{\text{grid}}_{t}$ the grid import power. The definitions of all variables and parameters are summarized in Table \ref{tab:nomenclature}.

Following Equation \ref{eq:prediction}, prediction enters the problem through $U^\text{pv}_t$, the per-step upper bound for the PV generation.
A predictive model $q_{\boldsymbol{\theta}}$ is trained to output $U^\text{pv}_t$, then the objective is fed into an optimizer to obtain the solution $\boldsymbol{x}^*$.
In this paper, we design a novel Markov Decision Process around $\boldsymbol{x}^*$ that closes the loop between prediction and optimization,  allowing efficient reinforcement learning algorithms to tackle the problem.



\section{Solving the Scheduling Problem with RL-PaO}
\label{sec:methodology}

Predicting the uncertain constraint in Equation \ref{eq:milp_full} shifts the objective problem and subsequently the solution schedule. Since the MILP optimizer internal states cannot be observed, conventional PtO/PaO methods assume that accurate prediction leads to optimized downstream costs.
However, we show the assumption is false.
Instead, we propose a novel MDP that permits an optimal policy to align prediction and  optimization to improve downstream cost in a principled manner.

\subsection{PtO and PaO}

Figure \ref{fig:framework}C describes the standard training paradigm for PtO and PaO, respectively.
PtO assumes that better estimation of the uncertain parameter  can lead to lower downstream operational costs.
However, this assumption rarely holds as the constraints can greatly influence the optimization objective and the resulting solution space. 
Recent studies show that there is no consistent correlation between prediction accuracy and downstream costs \citep{chenFeatureDrivenEconomicImprovement2022, mandi2024decision}.
Motivated by this, supervised PaO methods augment the MSE prediction loss by some function of cost, in a hope that SGD training could also lead to a balanced prediction and cost-awareness:
%
\begin{align*}
      \mathcal{L}_\text{PaO}(\boldsymbol{\theta}) = \alpha \cdot \underbrace{\widehat{\mathbb{E}}\left[ \left(q_{\boldsymbol{\theta}}(\boldsymbol{z}) - \boldsymbol{\xi}\right)^2 \right]}_\text{MSE} + \beta \cdot \underbrace{f(\boldsymbol{x}, \boldsymbol{\hat{\xi}})}_\text{Cost-aware},
\end{align*}
where $\alpha, \beta > 0$ are the weighting coefficients.
In existing literature, $\alpha, \beta$ are  typically computed through either heuristic search or a two-stage pre-training paradigm ~\citep{ gabriele2026forecasting,mandi2020smart, shah2022lodl}. 

The standard PaO formulation therefore has two intrinsic drawbacks:
(1) minimizing the augmented loss corresponds to learning multi-task behavior.  However, the conventional SGD can lead to solutions that neither minimizes the MSE loss nor achieving cost-awareness. 
(2) Naively balancing MSE with cost-awareness the term by a weighted sum often yields models that underperform the extreme cases: i.e. either $\alpha= 0$ or $\beta=0$.

\subsection{RL-PaO: a New RL paradigm for Solving the PaO Problem}


Our proposed method closes the loop by a novel Markov Decision Process in which the agent acts to predict $\boldsymbol{\xi}$ and receives a calibrated reward.\\[0.5em]
\textbf{Environment. } 
We deliberately include system modeling, optimizing, planning and execution into environment with which the agent interact. 
The RL agent observes current solution from the MILP optimizer as context to predict the unknown constraint for the next day-ahead schedule. 
The solution under goes execution and yields real cost as part of the reward.
This way, predictions can indeed be regarded as actions as they shift the environment (internal states of the MILP optimizer).
\\[0.5em]
\textbf{State. } The state includes sufficient information to predict the uncertain parameter $\boldsymbol{\xi}$ similar to other prediction methods. 
In our formulation, the continuous state space is $\mathbb{R}^{72}$, consisting of the optimal scheduling plan $\bm{x}^*$, which is a comprehensive operational profile (e.g., battery dispatch and grid purchasing) that implicitly reflects the latent meteorological patterns and temporal dynamics governing the PV upper bounds. 
To enable efficient learning, raw high-dimensional features are processed via dimensionality reduction before being fed into the agent.\\[0.5em]
\textbf{Action. } As indicated in the Problem Formulation, the predicted quantity is the per-step PV generation upper bound $\boldsymbol{\xi} := {U}_{t}^{\text{pv}}$ that resides in $\mathbb{R}^{22}$.
The actions are generated conditional on the dimensionality-reduced states.
Note that unlike the existing supervised PaO methods, our design naturally constructs a feedback loop since:
(1) prediction as action can influence the optimization and plan execution (through the inequality constraints)  due to the deliberate design, see Figure \ref{fig:framework};
(2) each action receives a reward designed so that maximizing cumulative reward improves the alignment between prediction and downstream optimization.
\\[0.5em]
\textbf{Transition. } After producing an action $U^\text{pv}_t$, it is used in the inequality constraint in Equation \ref{eq:milp_full}. We then invoke the MILP optimizer to solve the day-ahead scheduling problem to output the optimal schedule $\boldsymbol{x}^* \!\in\! \mathbb{R}^{72}$. 
Real-world execution then reveals the true PV output and the realized operational cost. 
Note that revealing $P^{\text{true}}_t$ is a long and highly complex process that does not readily permit online RL. Therefore, offline learning from the existing data logs is required.\\[0.5em]
\textbf{Reward. } Motivated by the observation that better prediction does not necessarily lead to better downstream cost performance~\citep{gabriele2026forecasting}, we design our reward to be an empirical mixture of prediction accuracy and decision quality measured by realized scheduling cost:
    \begin{equation}
    r_t = -\beta \cdot \texttt{N}(C_{\text{real}}) - (1-\beta) \cdot \texttt{N}(e_{\text{pred}})
    \end{equation}
    where $C_{\text{real}}$ is the actual electricity cost under the schedule, $e_{\text{pred}}$ is prediction error, and $\texttt{N}$ denotes normalization that scales the two reward terms to the same range.
    $\beta$ is a hyperparameter balancing decision quality and prediction stability.
Note that unlike the existing supervised PaO methods that augments loss, the association of cost to reward allows dynamically adjusting the agent behavior.

\textbf{Policy. }
It is well-known that an MDP permits an optimal policy if the states and actions are finite, rewards are bounded,  discount factor is less than one  \citep{Puterman1994}.
RL-PaO satisfies these assumptions and therefore  we are guaranteed that a well-calibrated reward function ideally can lead to optimal alignment between prediction and optimization to achieve lower downstream operational costs, provided the algorithm can attain global optimum.
This optimality stands in sheer contrast to the empirically designed mixture supervised loss of the conventional PaO methods~\citep{shah2022lodl, mandi2024decision} that may be stuck in local optima that minimize neither prediction error nor downstream cost surrogate.

\textbf{Training Algorithm.}
Having described the MDP, we use Proximal Policy Optimization (PPO) \citep{schulman2017proximal} to train the RL agent.
It has been shown that PPO can efficiently learn from offline data \citep{zhuang2023-behaviorPPO} and can converge to the global optimum \citep{Liu2019-NeuralPpoTrpoGlobalConvergence}. 
PPO iteratively updates the policy $\pi_\theta$ by maximizing the clipped surrogate objective:
\begin{equation}
L(\theta) = \mathbb{E}_t \left[ \min\left( r_t(\theta) A_t, \text{clip}(r_t(\theta), 1-\epsilon, 1+\epsilon) A_t \right) \right]
\end{equation}
where $r_t(\theta) = \frac{\pi_\theta(a_t|s_t)}{\pi_{\theta_{\text{old}}}(a_t|s_t)}$ is the probability ratio, $A_t$ is the advantage estimate, and $\epsilon$ controls the clipping range. 
The agent is trained in an offline fashion on historical data. In each training episode, the agent outputs an uncertain parameter estimate, then a black-box MILP optimizer is invoked to compute the corresponding optimal schedule.
Based on the schedule and the revealed cost,  reward is calculated against the realized true output. 
In the proposed framework, the MILP optimizer, its solution schedule plan and plan execution are all treated as part of the environment.

\section{Experiments}


We conduct experiments to answer the following research questions: 
\textbf{RQ1.} Is our proposed MDP well-posed in the sense that maximizing cumulative rewards equals minimizing downstream costs? 
\textbf{RQ2.} Can RL-PaO outperform the conventional PbO as well as PtO/PaO methods? 
\textbf{RQ3.} Can our method provide interpretability, including evidence that good prediction does not translate to better downstream operational costs? 
\textbf{RQ4.} Is our method robust towards hyperparameters?
\vspace{-3pt}

\subsection{Experimental Setup}

\textbf{Task.}
We use RL-PaO to tackle the day-ahead scheduling problem specified in Equation \ref{eq:milp_full}. 
We use a public dataset that records load, electricity price, temperature, etc. at the Osaka University in Japan\footnote{https://www.data.jma.go.jp/stats/etrn/}.
The training set spans 2016–2018 and the test set covers 2019, all at hourly resolution.
The MILP scheduling model is solved with the SCIP solver.
The objective is to minimize the total annual operational cost of the microgrid subject to  constraints.
A training episode traverses the entire training set sequentially, with one training step being one day.

\begin{figure*}[t!]
\centering
\includegraphics[width=.96\textwidth]{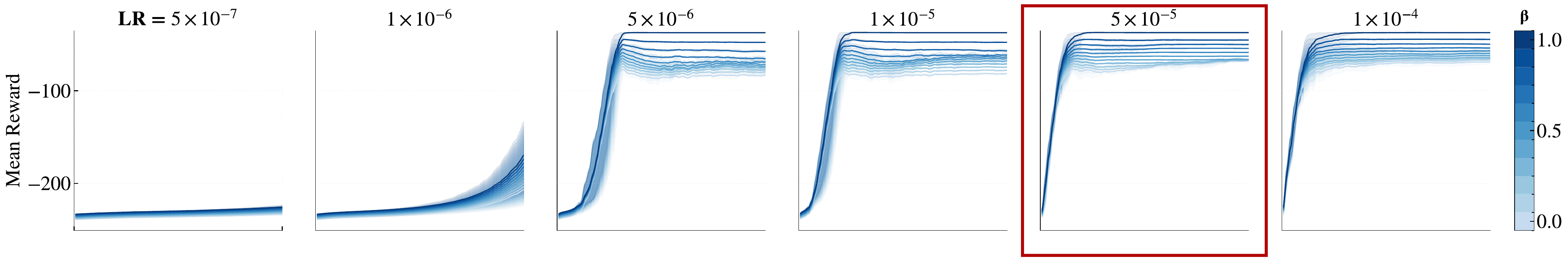}
\includegraphics[width=.96\textwidth]{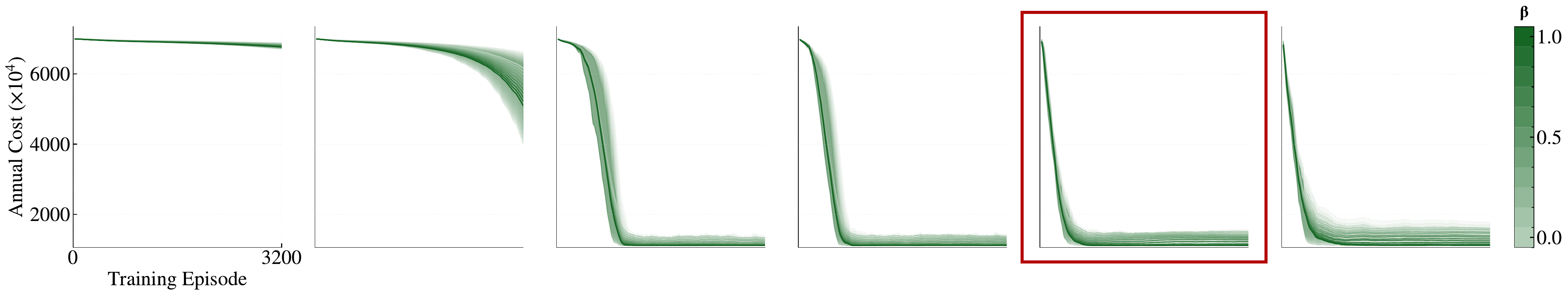}
\caption{
Cumulative rewards (upper row) and downstream operation costs (lower) across different learning rates.  Transparency of lines set according to $\beta$ value.
RL-PaO is effective in 
(1) establishing correspondence between high rewards and lower costs;
(2) the performance curves change smoothly along gradual increase of hyperparameters.
}
\label{fig:convergence}
\end{figure*}

\begin{figure}[!th]
\centering
\includegraphics[width=0.49\linewidth]{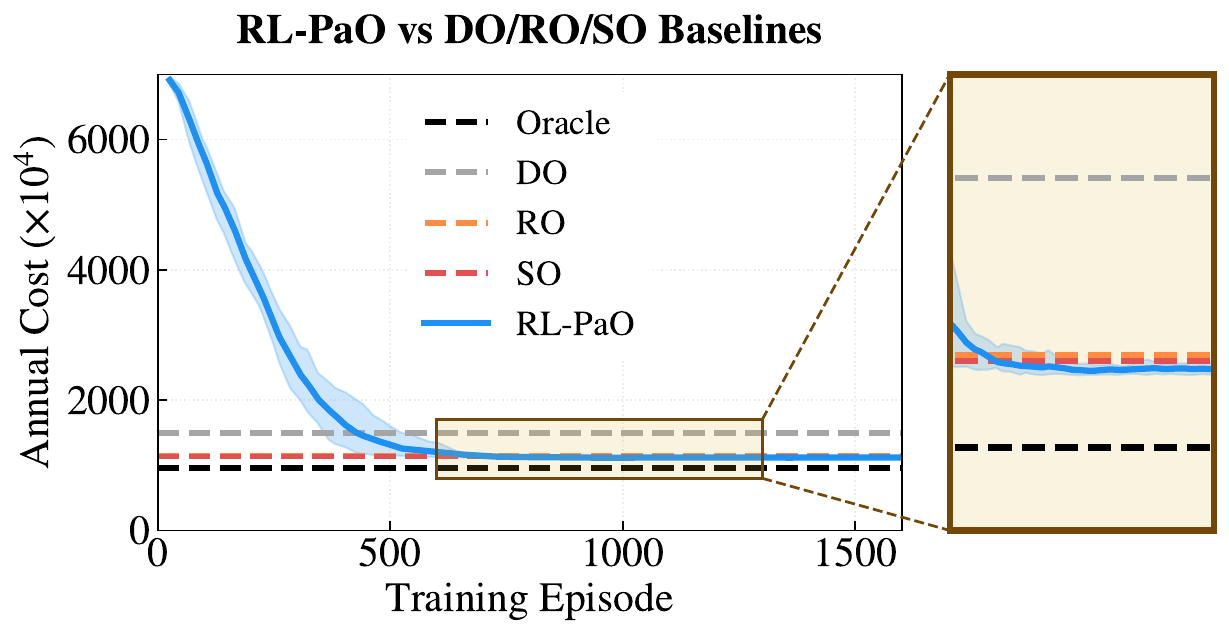}
\includegraphics[width=0.49\linewidth]{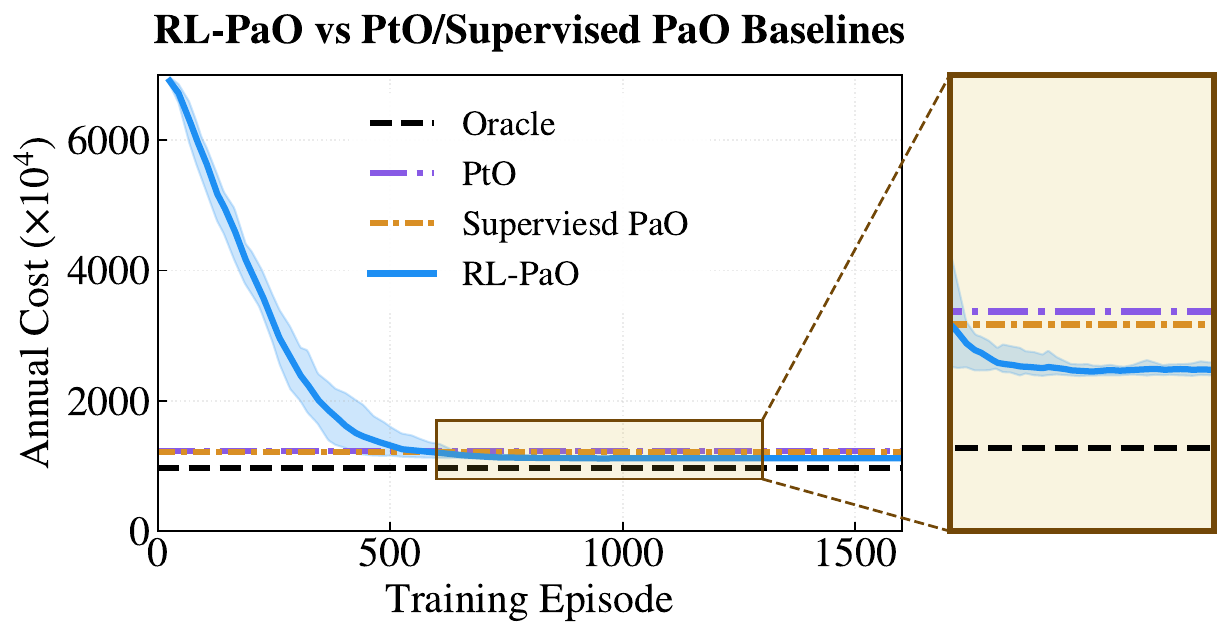}
\caption{
Comparison on annual cost.
Blue solid line and shade are mean and standard deviation over 5 independent runs.
 \textbf{Left}: RL-PaO compared against the conventional PbO baselines that require model knowledge in uncertainty (DO, RO, SO). 
 \textbf{Right}: comparison against PtO, PaO methods.
In both cases the model-free RL-PaO achieves the best cost.
}
\vspace{-5pt}
\label{fig:baseline_comparison}
\end{figure}

\textbf{Baselines.}
We select all three types of methods as baselines.
\emph{Oracle}: The full-information benchmark that solves the scheduling problem using ground-truth PV generation, load and electricity price, serving as the theoretical lower bound of operation cost.
\emph{PbO}: DO uses point forecast; 
RO is with the worst-case uncertainty sets, and SO is based on scenario generation.
\emph{PtO/PaO}: We follow the standard PtO practice to train using MSE prediction loss.
For PaO, we opt for the SPO+ surrogate regret loss~\citep{dupont2024decision}.

\subsection{Results}

\textbf{RQ1 \& RQ2: Performance. }
Figure \ref{fig:convergence} answers \textbf{RQ1} with an affirmative. 
It is visible that both cumulative rewards (upper) and downstream operational costs (lower) are improved along with training after learning rates are sufficient large, under all possible $\beta$ values.
This verifies that 
(1) our novel design that  exploits prediction as action to change downstream execution is well-posed, since converged policies can indeed achieve lower downstream costs;
(2) the RL-PaO formulation is smooth in hyperparameters in the sense that the performance curves change smoothly along with the increase of learning rates. There is no abrupt change after $\texttt{lr}\geq 5\times 10^{-6}$, though an excessively large value (e.g. $10^{-3}$) induces noticeable training oscillation. 
We consider $\texttt{lr} = 5\times10^{-5}, \beta=0.9$ to be the optimal configuration for subsequent analysis.

Regarding \textbf{RQ2},  Figure \ref{fig:baseline_comparison} shows that RL-PaO also outperforms the baselines in terms of annual cost.
The left hand side compares against the PtO baselines that require model knowledge on uncertainty.
Yet, it is visible in the zoom-in plot that model-free RL attains the best final annual cost, underperforming only the oracle.


In the RHS figure we see a similar trend. 
RL-PaO is also the best performer among PtO and supervised PaO, leading to around $9\times 10^5$ cost reduction than the second best method.
\begin{wraptable}[12]{r}{0.5\textwidth}
\centering
\vspace{-3.5pt}
\caption{Overall performance comparison on the 2019 test set.
Annual cost is measured in the unit of $\times 10^4$.
$\downarrow$ indicates lower is better.}
\vspace{-5pt}
\label{tab:performance}
\scriptsize
\setlength{\tabcolsep}{3pt}
\begin{tabular}{@{}llcc@{}}
\toprule
Category & Method
& Annual Cost $\downarrow$ & Cost Gap (\%) $\downarrow$ \\
\midrule
\multirow{4}{*}{PbO}
& Oracle & 963.76 & 0.00 \\
& DO & 1496.20 & 55.25 \\
& RO & 1145.20 & 18.83 \\
& SO & 1133.90 & 17.65 \\
\midrule
\multirow{3}{*}{PtO/PaO}
& PtO & 1233.40 & 27.98 \\
& Supervised PaO & 1207.90 & 25.33 \\
& \textbf{RL-PaO} & \textbf{1117.10} & \textbf{15.91} \\
\bottomrule
\end{tabular}
\end{wraptable}
This is because supervised PaO methods such as SPO+ require uncertain parameters to appear in the objective function, an assumption that is violated for the constraint-side PV uncertainty in our scheduling problem. RL-PaO naturally bypasses this limitation by directly optimizing end-to-end decision quality in a black-box manner.
Table \ref{tab:performance} further consolidates the observation: RL-PaO is the top player among all baseline methods.
Comparing to Oracle, RL-PaO achieves the lowest cost gap $15.91\%$ by model-free learning from data.

\begin{figure}[t]
\centering
\includegraphics[width=.92\textwidth]{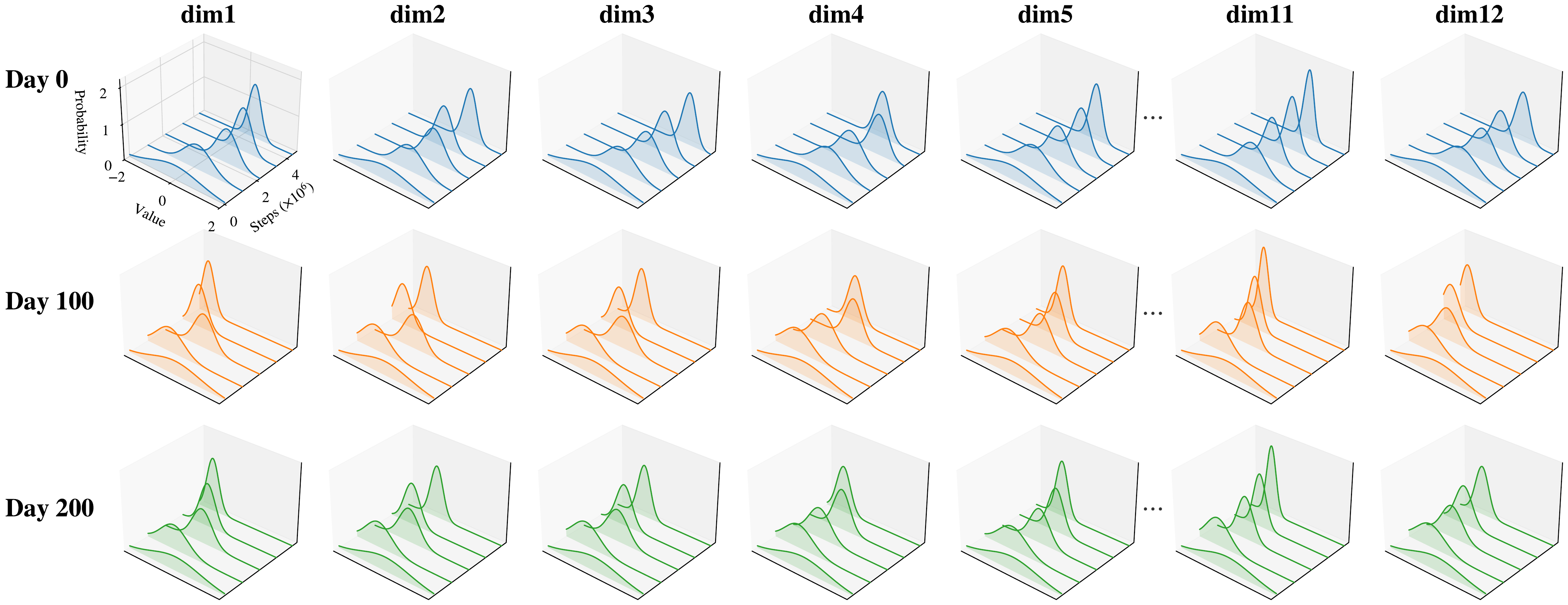}
\caption{
Policy evolution plot of the learned Gaussian distribution across 12 action dimensions at training day 0, 100, and 200.
Recall that prediction of the uncertain parameter is sampled from the policy.
As learning proceeds, the policies gradually narrow and shift toward the optimal decision region that aligns prediction with downstream cost optimization.
}
\label{fig:policy_evolution}
\end{figure}

\begin{figure*}[t]
\centering
\includegraphics[width=.92\textwidth]{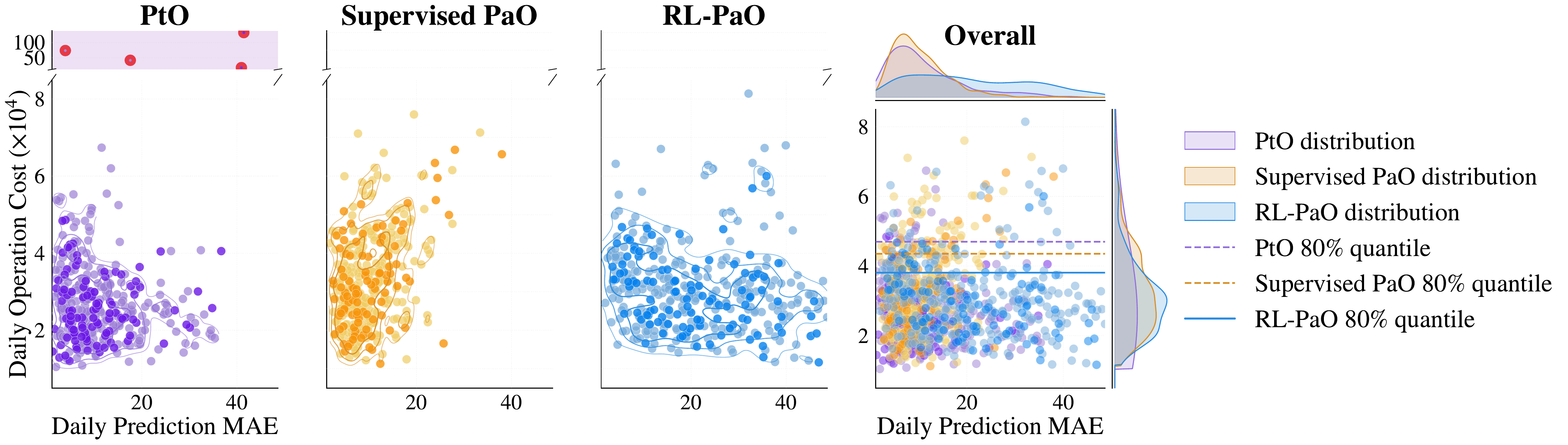}
\caption{
Prediction error versus operation cost, daily level.
The first three windows show scatter plots on the test set.
The last window shows a joint-distribution plot.
Transparency distinguishes weekdays and weekends.
PtO and PaO points are concentrated in a low estimation error region but the average cost is high. 
Outliers with extremely high costs are identified by red dots with purple background.
By contrast, RL-PaO error distribution is more dispersed, but the average cost is lower, as can be seen from the $80\%$ quantile line from the last window.
}
\vspace{-5pt}
\label{fig:joint_distribution}
\end{figure*}

\textbf{RQ3: Interpretability. }
Different from PtO/PaO baselines that do not provide interpretability for their predictions, RL-PaO offers insights from its policy space.
Figure \ref{fig:policy_evolution} shows policy evolution \citep{Zhu2025-qExpPolicy} of the learned Gaussian distribution across 12 action dimensions at training day 0, 100, and 200.
Prediction of the uncertain parameter is sampled from the policy.
Starting from a broad and nearly uniform distribution at day 0, the policy becomes progressively narrower and moves toward the optimal decision region that aligns prediction with downstream cost optimization.
The shift of the Gaussians at each dimension identifies different important regions that together characterize the optimal solution space for the studied problem.

\begin{figure*}[t]
\centering
\includegraphics[width=\textwidth]{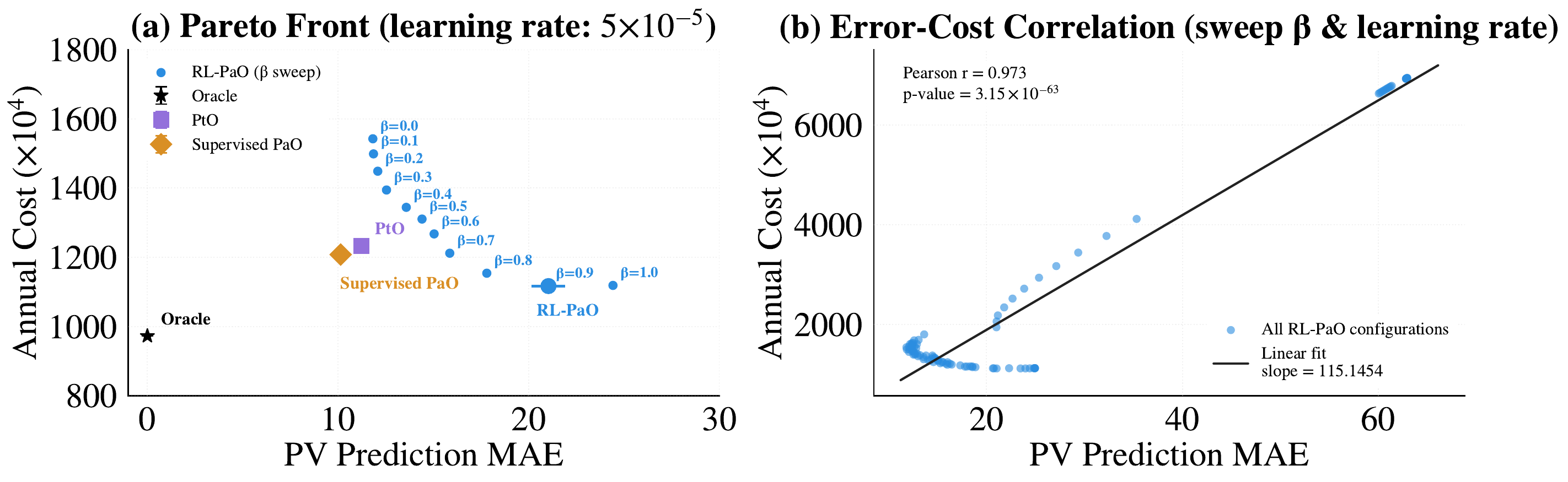}
\caption{Prediction error versus operation cost, annual level.
 (a) Pareto front of annual cost against prediction MAE, recorded for $\texttt{lr} = 5\times10^{-5}$. 
The reward weight $\beta$ spans a continuous trade-off curve and sits on the Pareto frontier, 
indicating that no competing methods can simultaneously achieve lower prediction error and lower operation cost.
 (b) Overall correlation across all $\beta$ and learning rate configurations.
 The blue points scattered in the lower left corner supports the finding that lower prediction error does not equal lower cost.
}
\label{fig:error_cost}
\end{figure*}

\begin{figure*}[t]
\centering
\includegraphics[width=\textwidth]{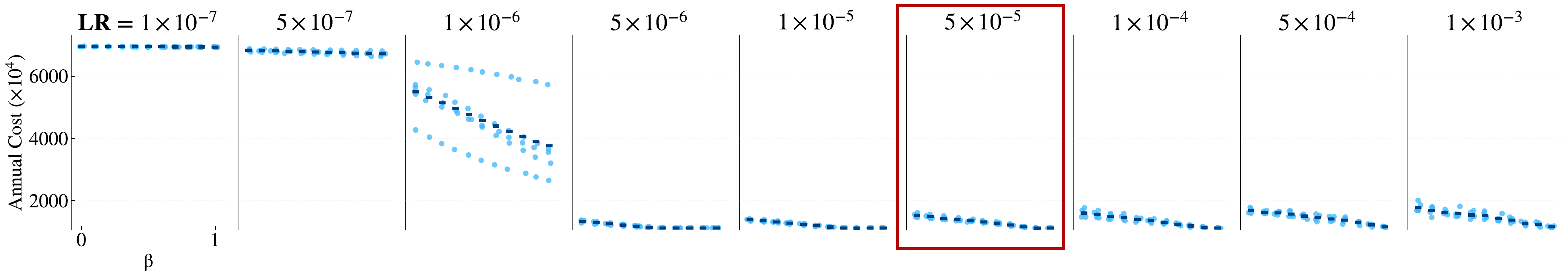}
\vspace{0.3em}
\includegraphics[width=\textwidth]{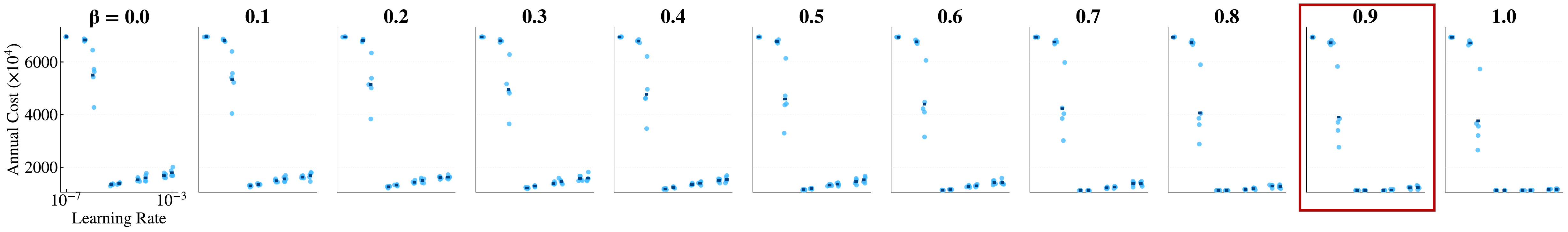}
\caption{Mahanttan plot of the proposed method across learning rates and reward weights $\beta$.
Transparent dots show final operational costs, with deep blue dots indicating the mean.
(Upper) Results of fixing $\beta$ and varying \texttt{lr}.
(Lower) Fixing $\texttt{lr}$ and varying $\beta$. 
The superior performance of RL can be attained from various combinations.
}
\label{fig:mahanttan_final_cost_lr_beta}
\end{figure*}

Fig.~\ref{fig:joint_distribution} visualizes the joint distribution of \emph{daily} prediction MAE and daily operation cost for four methods on the test dataset.
It is visible that PtO clusters in the region with low prediction error but relatively high operation cost, as its training objective purely minimizes fitting error without considering downstream scheduling impact. This observation also supports the finding that lower estimation errors does do not translate to lower costs.
It should also be noted that PtO has outliers shaded in purple, indicating extremely high costs are incurred -- a phenomenon unique to PtO.  We attribute this to its open-loop nature.
Supervised PaO shifts slightly toward lower cost but remains confined to a similar error-cost regime.
In contrast, RL-PaO disperses its error estimations but achieve significantly lower cost.
This is even more clear from the last joint distribution plot that estimates the densities of MAE and costs for all compared methods.
It can be seen that the RL-PaO's MAE distribution is more uniform, but its cost distribution places a majority of mass in lower-cost region, as indicated by the $80\%$ quantile line.
This observation further supports that high prediction accuracy does not equal better decision quality.


Figure \ref{fig:error_cost} further quantifies this relationship at the \emph{annual} level.
The LHS shows the Pareto front of prediction error versus annual total cost.
PtO achieves the lowest prediction error among the baselines but suffers from higher operation cost, lying in a Pareto-dominated region.
By adjusting the reward weight $\beta$, RL-PaO spans a continuous trade-off curve and sits on the Pareto frontier, 
indicating that no competing methods can simultaneously achieve lower prediction error and lower operation cost.
The RHS shows the overall correlation across all RL-PaO configurations with different $\beta$ and learning rates. 
The Pearson correlation coefficient is 0.973, indicating a strong overall positive relationship between prediction error and operation cost.
It is intuitive that for some of the configurations that lower MAE indeed indicates lower annual cost. But this relationship no longer holds for the majority of points scattered in the lower left corner.

\textbf{RQ4: Hyperparameter Sensitivity.} 
Figure \ref{fig:mahanttan_final_cost_lr_beta} shows a Mahanttan plot of the proposed method across learning rates and reward weights $\beta$.
Transparent dots are final operational costs of independent runs, with deep blue dots indicating the mean of the corresponding combinations.
The upper row shows results of fixing $\beta$ and varying \texttt{lr}, while the lower row fixing $\texttt{lr}$ and varying $\beta$. 
It is visible that the superior performance of the proposed method is not a result of specific hyperparameter setting but can be attained from various combinations.




\section{Conclusion}

Predicting the parameters is crucial for decision making under uncertainty.
PtO  and PaO predict the parameter through independent prediction loss and decision-augmented surrogate, respectively.
This paper presented RL-PaO, an RL framework that treats prediction as action and integrates system formulation, optimization, and decision execution into a single environment, yielding an MDP that could be solved effectively by existing RL methods.
Experiments on the day-ahead scheduling problem using real data at the Osaka University verified that RL-PaO achieved lowest annual cost than all non-oracle baselines.  
Extensive analysis showed that our RL-PaO design can also provide strong interpretability and is robust towards hyperparameters.

We identify a potential limitation for this work: while the black-box treatment of the downstream problem delivers strong generalizability, it also introduces computational overhead during training.  Every training step requires solving the full downstream optimization problem and runtime can become prohibitive for large-scale complex systems with a vast set of decision variables. 

Future work can be pursued in two directions. 
First, it would be helpful to extend to joint prediction of multiple uncertain parameters of different physical natures, including load, electricity prices and distributed renewable output, and examine the performance of a single RL agent in such multi-variable settings. Second, generating longer-horizon schedules such as one-week dispatch plans could also help alleviate computational burden.

\bibliography{iclr2027_conference}
\bibliographystyle{iclr2027_conference}

\clearpage
\appendix
\section{Appendix}

\subsection{Related Work}

\textbf{PbO.} DO replaces uncertainties parameters with deterministic estimates~\citep{boyd2004convex, bertsekas1999nonlinear} but loses feasibility and optimality under time-varying uncertainty. RO guarantees worst-case feasibility over a predefined uncertainty set~\citep{bental2002robust, bertsimas2004robust} but yields over-conservative solutions with excessive costs. SO optimizes expected cost via probability distributions~\citep{shapiro2009lectures, birge2011stochastic}, yet relies on valid distributions, neglects action risks, and is computationally expensive at scale.

\textbf{PtO.} PtO decouples prediction and optimization: a predictor minimizes forecast error, and a solver generates decisions from predicted parameters~\citep{elmachtoub2022spo}. This modular framework and allows independent module tuning, with applications spanning microgrid dispatch~\citep{wen2019microgrid}, inventory management~\citep{gallien2015zara}, and retail pricing. \citep{ferreira2016analytics} combined random forest forecasts with integer programming for pricing; \citep{gallien2015zara} paired linear regression with MILP for inventory optimization. In energy systems ~\citep{zhao2023data, chenFeatureDrivenEconomicImprovement2022}, LSTMs predict renewable generation for downstream solvers. However, PtO suffers from prediction-decision mismatch, and lower forecast error does not equal better decision.

\textbf{PaO.} To address prediction-decision mismatch, PaO optimizes directly for downstream cost rather than pure prediction error~\citep{kotary2021end,lahoud2025predict,mandi2024decision}. Gradient-based differentiable PaO embeds solvers as differentiable layers for end-to-end training, including KKT-based OptNet~\citep{amos2017optnet, donti2017task, ferber2020mipaal} and interior-point regularized methods~\citep{mandi2020interior}; both require problem-specific derivation and are sensitive to downstream model complexity. Surrogate loss and perturbation approaches avoid explicit differentiation: SPO constructs surrogate losses via duality theory~\citep{elmachtoub2022spo}, and perturbation-based schemes estimate gradients from solver outputs for black-box compatibility~\citep{berthet2020learning, pogancic2020differentiation, shah2022lodl}, trading generality for reduced gradient fidelity. Non-gradient discrete PaO bypasses gradient computation and optimizes decision loss directly~\citep{elmachtoub2020decision, ban2019big}, but is limited to simple architectures and lacks high-dimensional scalability. Engineering techniques including warm-starting, approximate solvers, and pre-computed surrogates cut PaO training cost~\citep{mandi2020smart, wang2020automatically, kong2022end}. More recently, \citep{mandi2025feasibility} propose losses that balance infeasibility and suboptimality without restricting the downstream problem to LP/MIP.

\subsection{Implementation Details}
\label{app:implementation_details}

\begin{table}[htbp]
\centering
\caption{Hyperparameters and configurations.}
\label{tab:hyper_params}
\resizebox{0.825\linewidth}{!}{
\begin{tabular}{lc}
\hline
Hyperparameter & Value \\
\hline
Reward weight $\beta$ & Swept in $\{0.0, 0.1, 0.2, \dots, 1.0\}$ \\
Learning rate & \begin{tabular}[c]{@{}l@{}}Swept in $\{10^{-7},\, 5{\times}10^{-7},\, 10^{-6},\, 5{\times}10^{-6},$\\ $10^{-5},\, 5{\times}10^{-5},\, 10^{-4},\, 5{\times}10^{-4},\, 10^{-3}\}$\end{tabular} \\
Discount factor $\gamma$ & 0.99 \\
PPO clip range & 0.2 \\
GAE parameter $\lambda$ & 0.95 \\
Value function coefficient $v_f$ & 0.5 \\
Max gradient norm & 0.5 \\
Epochs per policy update & 10 \\
Rollout steps per update & 2048 \\
Mini-batch size & 64 \\
Hidden layer size (actor / critic) & 64 \\
Number of hidden layers (actor / critic) & 2 \\
Base random seed & 42 \\
Number of independent random seeds & 5 \\
Steps per training episode & 1096 \\
MILP relative optimality gap & 0.01 \\
MILP time limit per solve & 0.3\,s \\
\hline
\end{tabular}
}
\end{table}

\noindent \textbf{Dataset and Environment.}
The electricity price, solar irradiance, ambient temperature data etc. are sourced from public datasets.\footnote{https://www.data.jma.go.jp/stats/etrn/}
The training set spans years 2016--2018, and the held-out test set covers year 2019. All data are at 2-hour resolution, yielding 12 time steps per day for the day-ahead scheduling horizon.

\noindent \textbf{Optimization Model.}
The day-ahead energy scheduling problem is formulated as a mixed-integer linear program (MILP). The objective minimizes total operational cost. The model is solved via the SCIP solver, with a 1\% relative optimality gap and a 0.3\,s time limit per solve to ensure computational efficiency during reinforcement learning training.

\noindent \textbf{RL-PaO Agent.}
The RL agent is built on the Proximal Policy Optimization (PPO) algorithm with a standard actor-critic MLP architecture. Raw high-dimensional scheduling solution vectors from the MILP solver are compressed into low-dimensional observations via principal component analysis (PCA). The composite reward function is defined as
$r = -\beta \cdot J_{\text{cost}} - (1-\beta) \cdot J_{\text{MSE}}$,
where $\beta \in [0,1]$ balances the operational cost term and the prediction error term.

\noindent \textbf{Training and Hyperparameter Sweep.}
We sweep the reward weight $\beta \in [0,1]$ and the policy learning rate $\text{lr} \in [10^{-7}, 10^{-3}]$. All results are averaged over 5 independent random-seed runs, with min--max ranges shown as shaded areas in the figures. The full framework is implemented in Python. Each training step corresponds to one day-ahead scheduling instance. One full training episode comprises 1096 steps, forming a complete sequential pass over the entire 2016–2018 training set.

\subsection{Additional Results}
This section provides supplementary daily cost breakdowns and detailed ablation analyses to support the findings presented in the main paper.

\setcounter{figure}{0}
\renewcommand{\thefigure}{C\arabic{figure}}

\begin{figure*}[htbp]
\centering
\includegraphics[width=\textwidth]{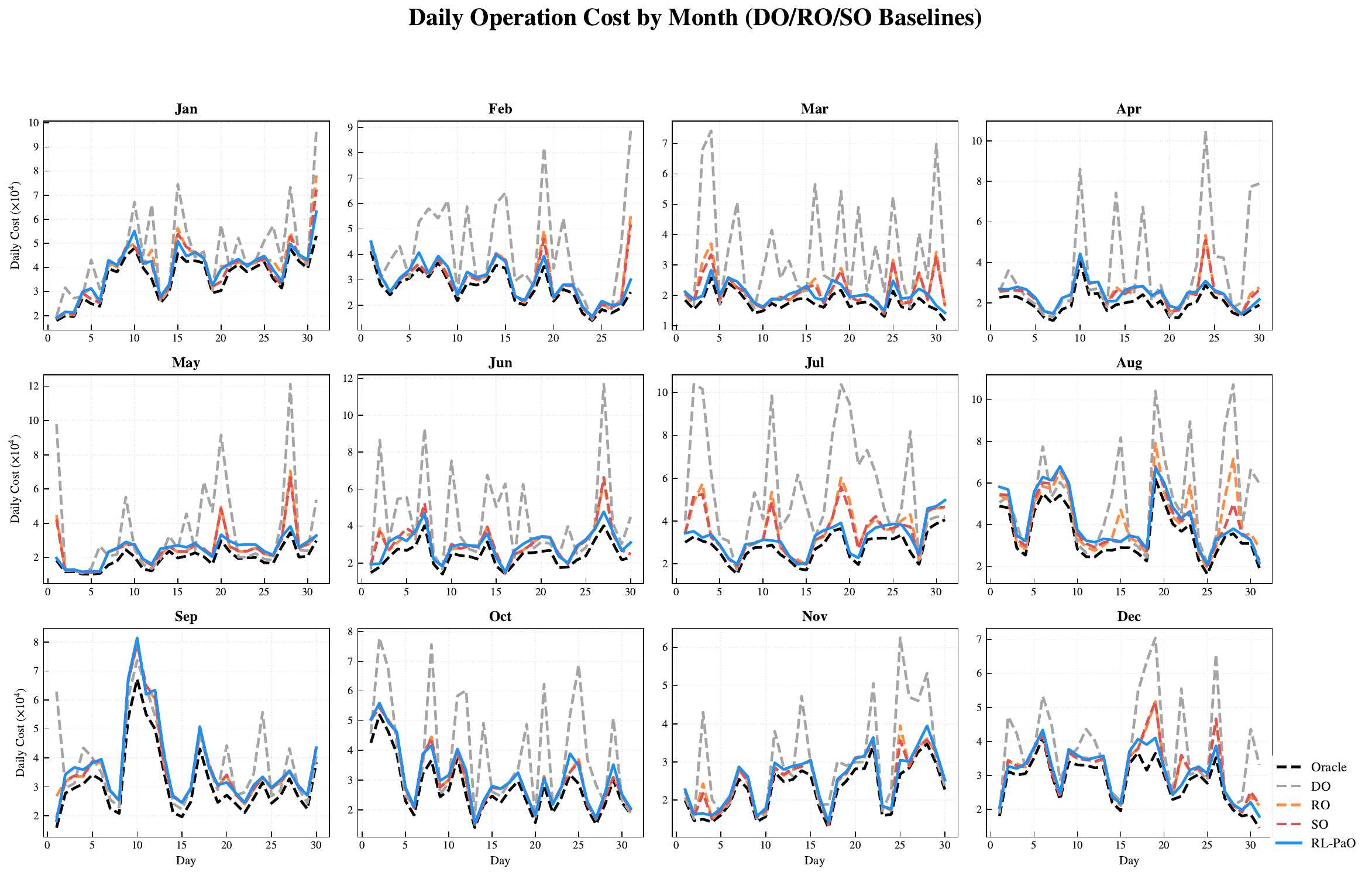}
\caption{Daily operation cost by month compared with conventional optimization baselines on the 2019 test set. Each subplot corresponds to one calendar month, comparing Oracle, DO, RO, SO and the proposed RL-PaO.}
\label{fig:monthly_optimization}
\end{figure*}
Figure C1 decomposes the daily operational cost on a per-month basis over the 2019 test set, comparing the proposed RL-PaO with three conventional DO, RO, and SO, as well as the full-information Oracle lower bound. The DO baseline shows the highest cost volatility and overall cost level, with particularly large deviations. RO and SO deliver more stable performance but still remain significantly above the Oracle benchmark. By contrast, RL-PaO closely follows the Oracle costs across all twelve months, maintaining near-ideal performance under diverse seasonal PV generation conditions.

\begin{figure*}[htbp]
\centering
\includegraphics[width=\textwidth]{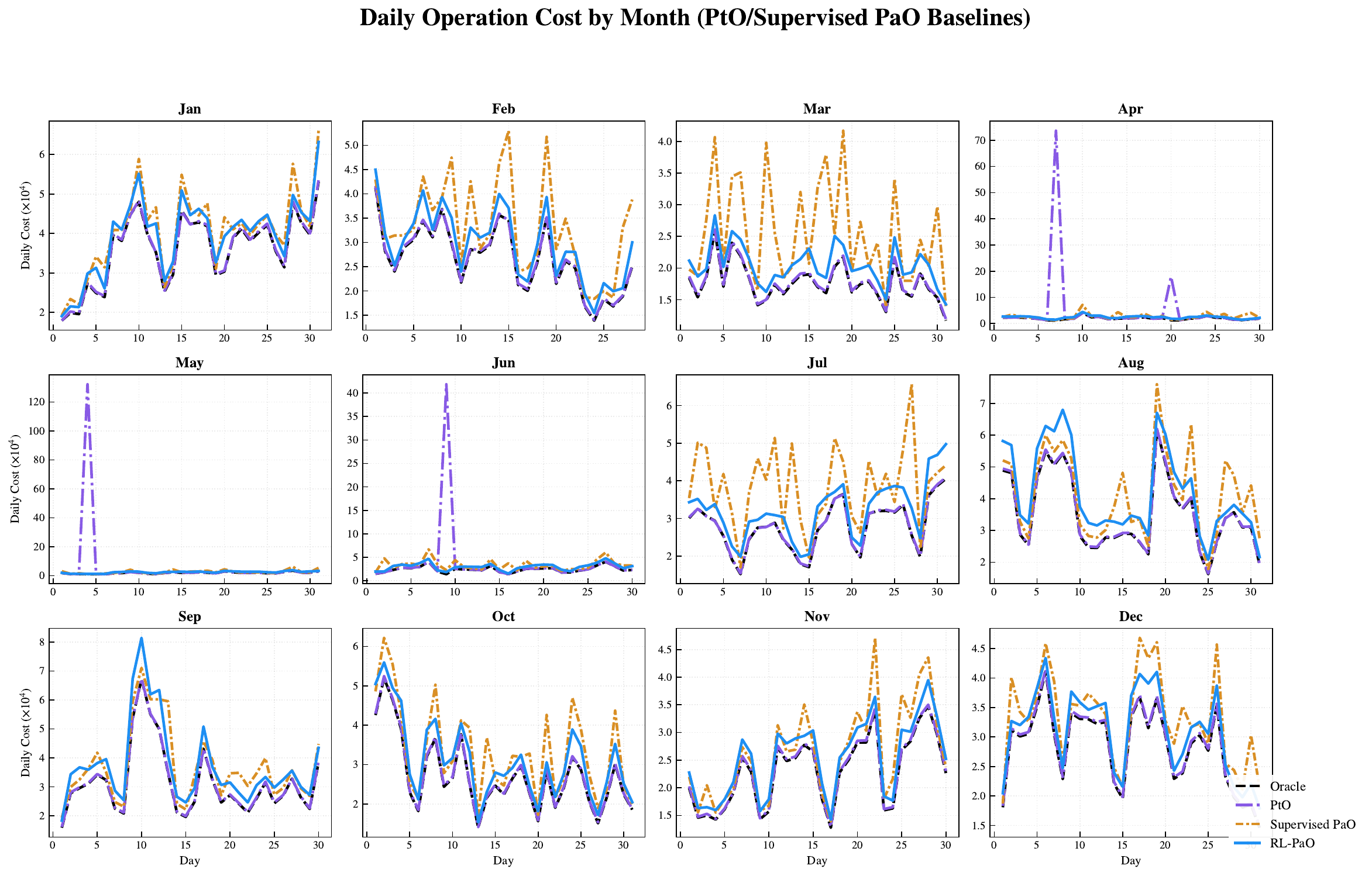}
\caption{Daily operation cost by month compared with predict-and-optimize baselines on the 2019 test set. Each subplot corresponds to one calendar month, comparing Oracle, predict-then-optimize (PtO), supervised PaO, improved PaO and the proposed RL-PaO.}
\label{fig:monthly_prediction}
\end{figure*}
Figure C2 presents the monthly daily cost comparison against P-O baselines, including the standard two-stage PtO framework and supervised PaO. The PtO approach exhibits extreme cost spikes in spring and summer months (April–June), reflecting the unstableness of pure prediction-error-driven training under constraint uncertainty. Supervised PaO improves upon PtO but still suffers from notable performance fluctuations. The proposed RL-PaO achieves cost levels closest to the Oracle across all months without anomalous peaks, demonstrating substantially stronger robustness than both supervised prediction-optimization baselines.


\begin{figure*}[htbp]
\centering
\includegraphics[width=\textwidth]{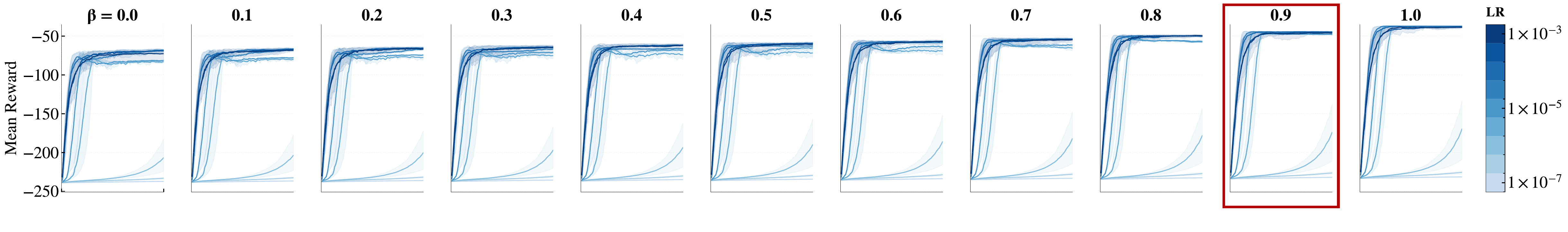}
\vspace{0.3em}
\includegraphics[width=\textwidth]{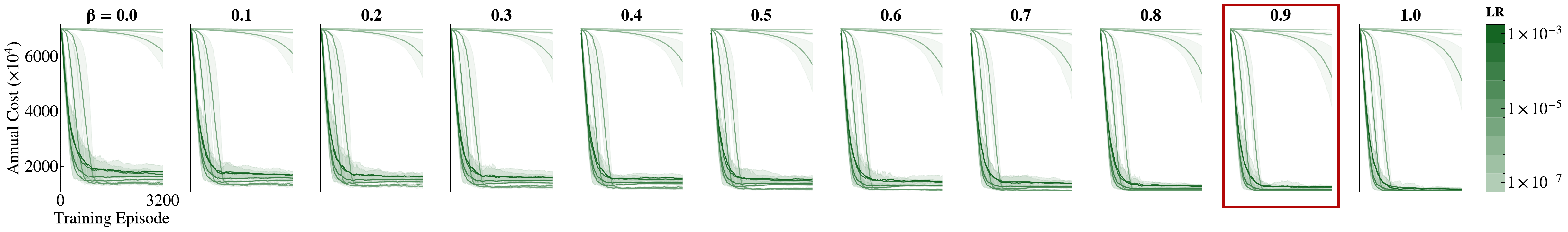}
\caption{Convergence behavior of RL-PaO across different reward weights $\beta$ and learning rates. Top: mean reward convergence across training episodes. Bottom: annual operation cost convergence across training episodes. Each column corresponds to a fixed $\beta$ value, and each colored curve corresponds to a different learning rate.}
\label{fig:appendix_convergence_beta_lr}
\end{figure*}

Figure C3 presents the full convergence behavior of RL-PaO across a grid of reward weights $\beta$ and learning rates, with mean reward shown in the top row and annual operation cost in the bottom row. Learning rate dominates convergence speed: configurations with very low learning rates fail to converge within the training horizon, while moderate to high rates converge within 500 episodes. Reward weighting $\beta$ determines the final performance level: higher $\beta$ consistently yields better final reward and lower operational cost across all learning rates. This figure is also presented and discussed in the main experimental section.


\begin{figure*}[htbp]
\centering
\includegraphics[width=0.48\textwidth]{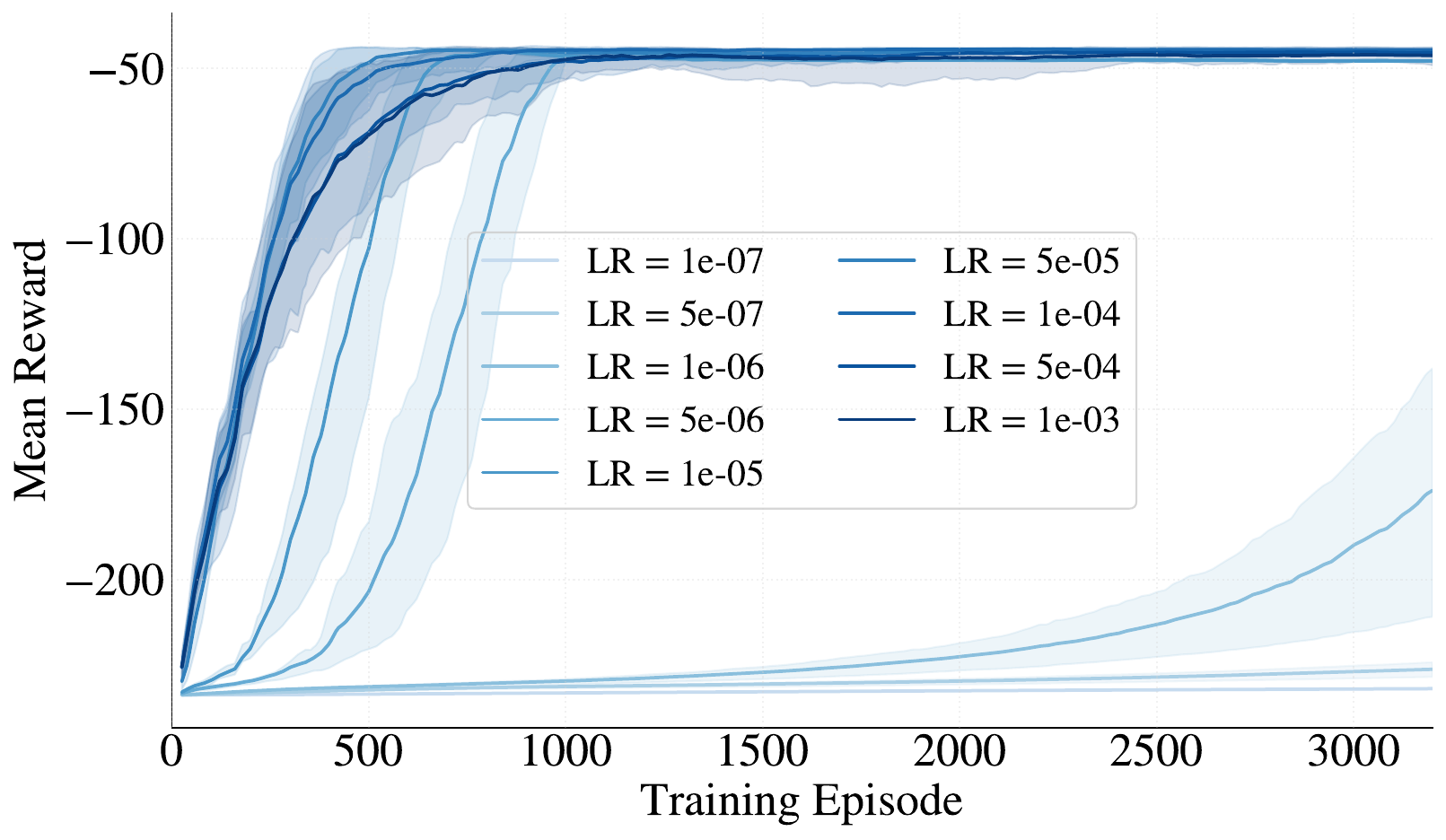}
\hfill
\includegraphics[width=0.48\textwidth]{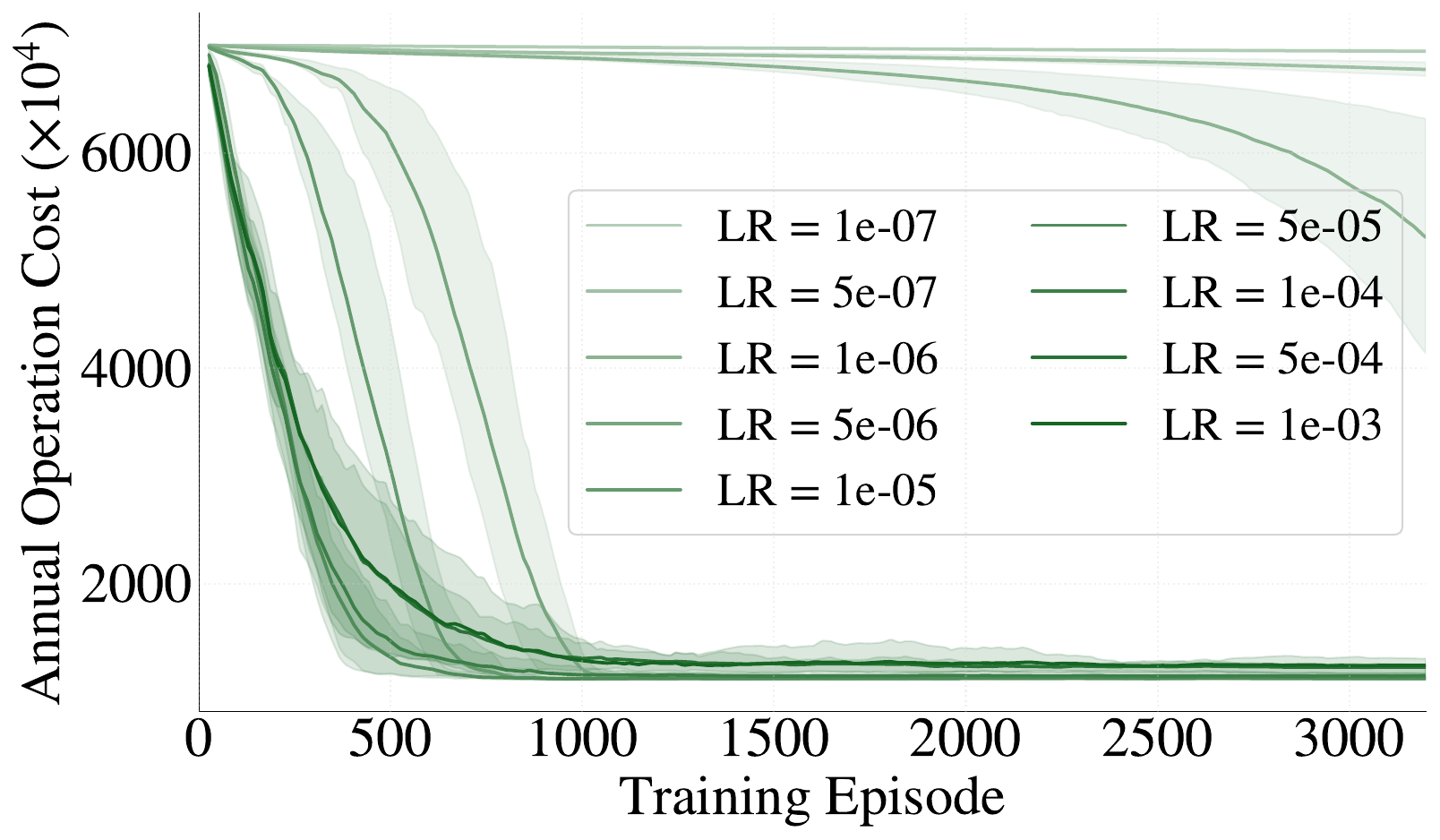}
\caption{\textbf{Left:} Reward and operation cost convergence under different learning rates with $\beta$ fixed at 0.90. Each colored band corresponds to one learning rate configuration. \textbf{Right:} Reward convergence under different reward weights $\beta$ with learning rate fixed at $5\times10^{-5}$. Each colored band corresponds to one $\beta$ configuration.}
\label{fig:appendix_ablation_lr_reward}
\end{figure*}

Figures C4 present mean reward convergence and annual operation cost convergence, respectively, across different learning rates with $\beta$ fixed at 0.90. Learning rate is the dominant factor governing convergence speed: very low learning rates (e.g., $10^{-7}$) converge extremely slowly and fail to approach the optimal performance level even after 3000 training episodes. Convergence accelerates markedly as the learning rate increases, with moderate rates in the $5\times10^{-5}$ to $10^{-4}$ range reaching a stable performance plateau within approximately 500 episodes. In terms of final solution quality, moderate learning rates yield the highest mean reward (Figure C4) and the lowest operational cost (Figure C5) with the tightest variance across random seeds. Excessively high learning rates (e.g., $10^{-3}$) do not further speed up convergence, but instead cause marginal performance degradation and larger fluctuation, as overly large parameter update steps tend to overshoot the optimal policy. This jointly confirms that the learning rate imposes a trade-off between convergence speed and final scheduling performance, and the $5\times10^{-5}$ setting used in the main experiment strikes the optimal balance between fast convergence, near-optimal cost, and training stability.

\begin{figure*}[htbp]
\centering
\includegraphics[width=0.48\textwidth]{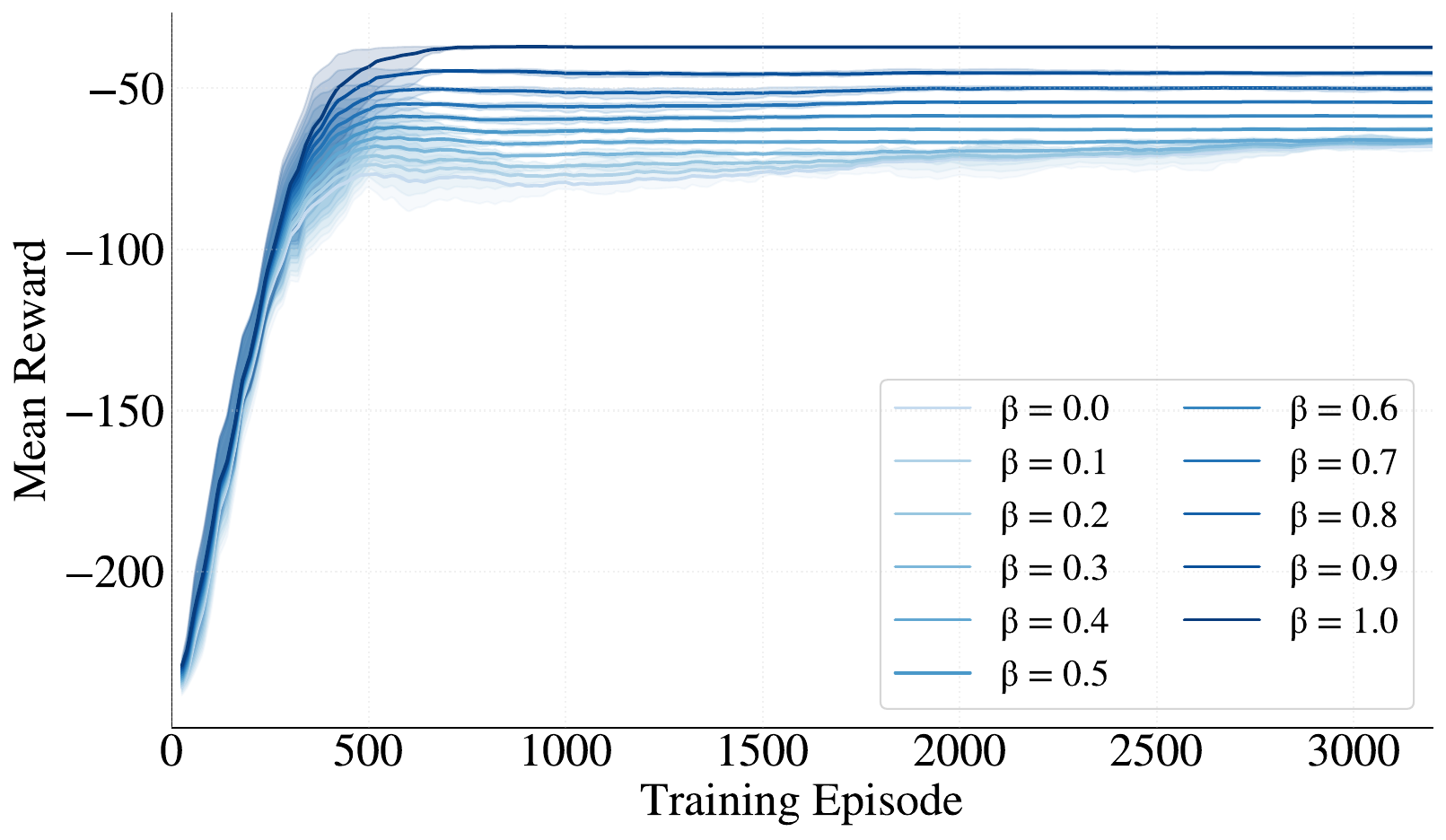}
\includegraphics[width=0.48\textwidth]{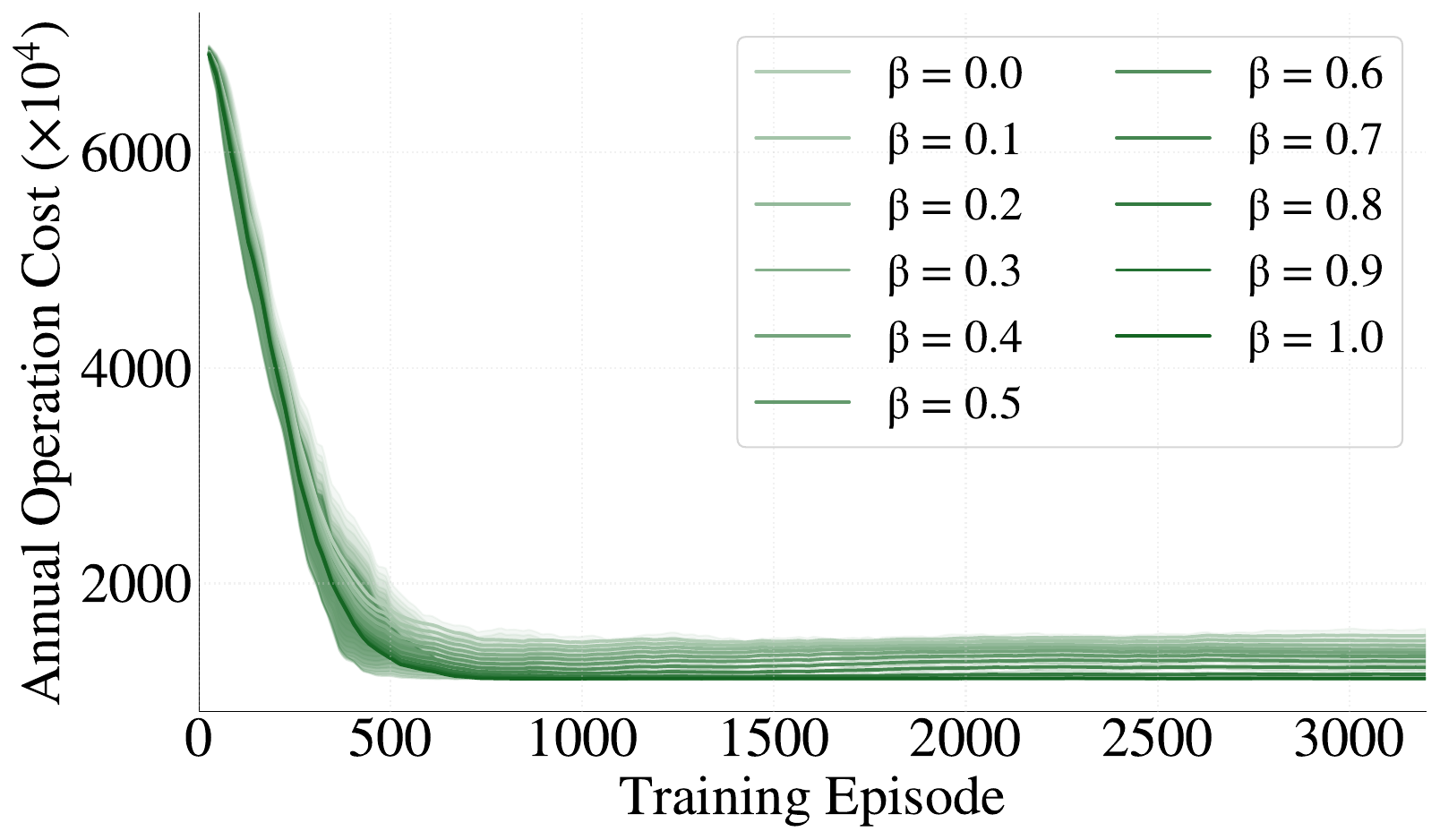}
\caption{\textbf{Left:} Annual operation cost convergence under different reward weights $\beta$ with learning rate fixed at $5\times10^{-5}$. Each colored band corresponds to one $\beta$ configuration. \textbf{Right:} Annual operation cost convergence under different learning rates with $\beta$ fixed at 0.90. Each colored band corresponds to one learning rate configuration.}
\label{fig:appendix_ablation_beta_reward}
\end{figure*}

﻿
Figures C5 present reward convergence and annual operation cost convergence, respectively, across different reward weights $\beta$ with the learning rate fixed at $5\times10^{-5}$. All $\beta$ configurations exhibit similar convergence speeds, reaching a stable performance plateau around 500 training episodes, indicating that the cost weighting coefficient barely affects the convergence rate of the RL training pipeline. In terms of final performance, there is a consistent monotonic trend: higher $\beta$ values yield higher mean reward (Figure C6) and lower operational cost (Figure C7) as the model places greater emphasis on minimizing downstream scheduling cost. Notably, the performance gain diminishes as $\beta$ approaches 1.0: $\beta=0.9$ already achieves near-optimal cost levels comparable to $\beta=1.0$, while maintaining tighter variance across random seeds. This jointly confirms that cost-oriented reward weighting effectively improves decision quality, and $\beta=0.9$ strikes the optimal balance between final scheduling performance and training stability.

\begin{figure*}[htbp]
\centering
\includegraphics[width=0.85\textwidth]{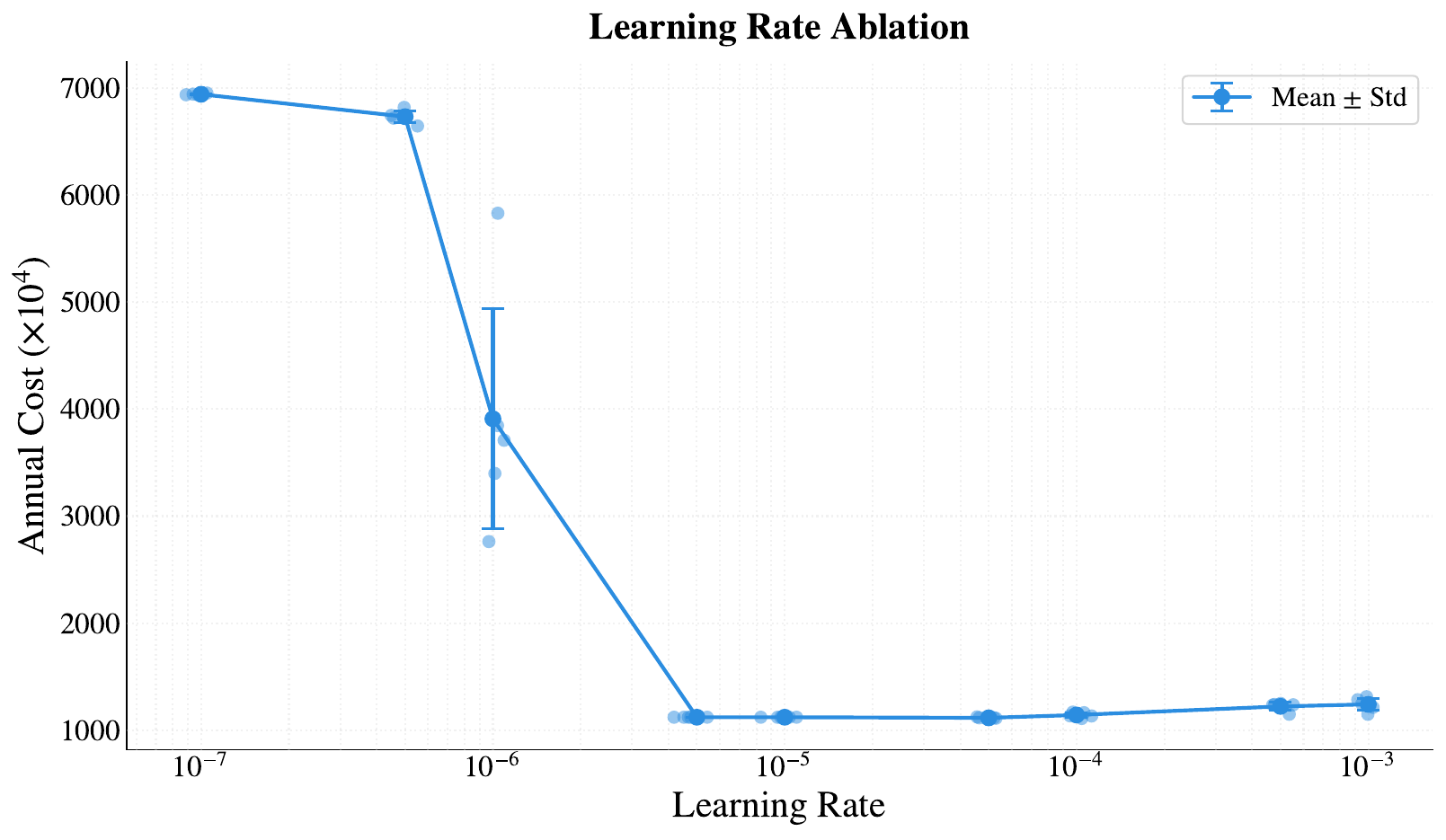}
\caption{Final annual operation cost with mean and standard deviation across different learning rates, with $\beta$ fixed at 0.90.}
\label{fig:appendix_ablation_lr_final}
\end{figure*}
Figure C6 quantifies the final annual operational cost (mean and standard deviation) across different learning rates with $\beta$ fixed at 0.90. The cost stays at an extremely high level when the learning rate falls below $10^{-6}$, as the algorithm fails to converge sufficiently within the training horizon. It drops to the lowest plateau across the $10^{-5}$ to $10^{-4}$ range, and climbs slowly as the learning rate increases further past $5\times10^{-5}$. This pattern indicates that the learning rate is not the higher the better: excessively high learning rates produce overly large parameter update steps that tend to overshoot the optimal solution and induce training oscillation, which leads to marginally suboptimal final scheduling performance and degraded stability. The $5\times10^{-5}$ setting adopted in the main experiment achieves the balance between convergence speed and solution quality.

\begin{figure*}[htbp]
\centering
\includegraphics[width=0.85\textwidth]{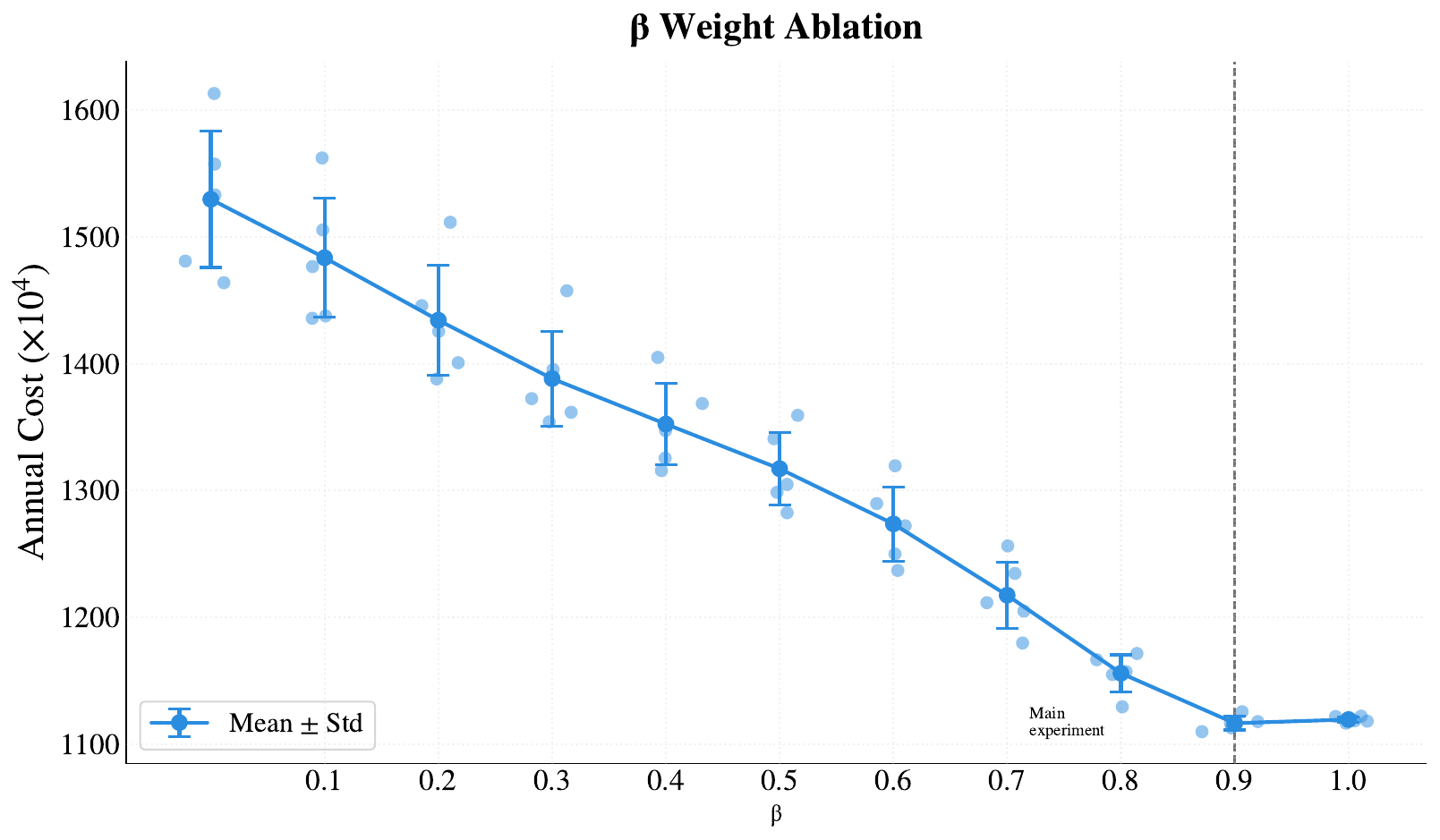}
\caption{Final annual operation cost with mean and standard deviation across different reward weights $\beta$, with learning rate fixed at $5\times10^{-5}$. The dashed line marks the $\beta$ value adopted in the main experiment.}
\label{fig:appendix_ablation_beta_final}
\end{figure*}

Figure C7 quantifies the final annual operational cost (mean and standard deviation) across different reward weights $\beta$ with the learning rate fixed at $5\times10^{-5}$. Overall, the cost decreases monotonically as $\beta$ increases from 0.0 to 0.9, reaching the minimum (optimal) level at $\beta = 0.9$. However, the cost rebounds slightly when $\beta$ further increases to 1.0. This pattern indicates that excessively weighting the operational cost term while compromising the reward regularization component degrades training stability and leads to marginally suboptimal final performance. The dashed line marks the $\beta = 0.9$ value adopted in the main experiment, which achieves the balance between cost-oriented decision optimization and training stability.

\end{document}